\pdfoutput=1

\documentclass[11pt]{article}

\usepackage[preprint]{acl}

\usepackage{times}
\usepackage{latexsym}

\usepackage{algorithm}

\usepackage{graphicx}%
\usepackage{multirow}%
\usepackage{amsmath,amssymb,amsfonts}%

\usepackage{amsthm}%
\usepackage{mathrsfs}%
\usepackage[title]{appendix}%
\usepackage{xcolor}%
\usepackage{textcomp}%
\usepackage{manyfoot}%
\usepackage{booktabs}%
\usepackage{algpseudocode}%
\usepackage{listings}%
\usepackage{anyfontsize}
\usepackage{color}
\usepackage{bbding}
\usepackage{longtable}
\usepackage{supertabular}
\usepackage{colortbl}
\usepackage[T1]{fontenc}

\usepackage[utf8]{inputenc}

\usepackage{microtype}

\usepackage{inconsolata}

\usepackage{graphicx}

\title{MAP: Revisiting Weight Decomposition for Low-Rank Adaptation}

\author{Chongjie Si$^1$, Zhiyi Shi$^2$, Yadao Wang$^3$, Xiaokang Yang$^1$, Susanto Rahardja$^4$, Wei Shen$^1$ \\
 $^1$Shanghai Jiao Tong University, $^2$Harvard University\\ $^3$Alibaba Group, $^4$Singapore Institute of Technology\\
  \texttt{\{chongjiesi, wei.shen\}}@sjtu.edu.cn \\}

\begin{document}
\maketitle

\begin{abstract}

The rapid development of large language models has revolutionized natural language processing, but their fine-tuning remains computationally expensive, hindering broad deployment. 
Parameter-efficient fine-tuning (PEFT) methods, such as LoRA, have emerged as solutions.
Recent work like DoRA attempts to further decompose weight adaptation into direction and magnitude components. 
However, existing formulations often define direction heuristically at the column level, lacking a principled geometric foundation.
In this paper, we propose MAP, a novel framework that reformulates weight matrices as high-dimensional vectors and decouples their adaptation into direction and magnitude in a rigorous manner.
MAP normalizes the pre-trained weights, learns a directional update, and introduces two scalar coefficients to independently scale the magnitude of the base and update vectors. This design enables more interpretable and flexible adaptation, and can be seamlessly integrated into existing PEFT methods. 
Extensive experiments show that MAP significantly improves performance when coupling with existing methods, offering a simple yet powerful enhancement to existing PEFT methods.
Given the universality and simplicity of MAP, we hope it can serve as a default setting for designing future PEFT methods.
\end{abstract}

\section{Introduction}

The rise of large-scale pre-trained language models, such as GPT-3 \cite{brown2020language}, BERT \cite{devlin2018bert}, and RoBERTa \cite{liu2019roberta}, has led to transformative advancements in natural language processing. 
These models have achieved remarkable success in tasks ranging from task-specific adaptation \cite{luo2023wizardmath, yu2023metamath} to instruction-following \cite{ouyang2022training} and aligning with human preferences \cite{bai2022training, rafailov2024direct}. 
Despite their impressive capabilities, fine-tuning these models, which often contain hundreds of millions to billions of parameters, remains computationally expensive, presenting significant obstacles to their widespread deployment \cite{raffel2020exploring, qiu2020pre}.

To mitigate this challenge, parameter-efficient fine-tuning (PEFT) has emerged as a promising solution \cite{zhang2022adaptive, si2025unleashing, houlsby2019parameter}, focusing on optimizing a small subset of model parameters to achieve high task performance while maintaining the integrity of the pre-trained model. 
Among the various PEFT techniques, Low-Rank Adaptation (LoRA) \cite{hu2021lora} has become a widely adopted approach.
LoRA updates the frozen weights $\mathbf{W}$ by adding a low-rank update matrix $\Delta \mathbf{W}$, leading to fine-tuned weights expressed as $\mathbf{W} + \Delta \mathbf{W}$. 
It has demonstrated both computational efficiency and scalability, and has inspired a range of subsequent methods that build on the low-rank adaptation framework \cite{si2024flora, zhang2022adaptive}.

Recently, DoRA \cite{liu2024dora} has been proposed to decouple the magnitude and direction of weight adaptation during fine-tuning. 
Specifically, DoRA normalizes the sum of the pre-trained weight and the low-rank update, $\mathbf{W} + \Delta\mathbf{W}$, in a per-column fashion and then rescales each column with a learnable vector. 
While DoRA introduces a novel perspective, it also exhibits a key limitation: it defines the ``direction'' of a matrix through column-wise normalization.
However, it remains unclear why the notion of matrix direction should be interpreted on a per-column basis, rather than alternatives such as row-wise normalization—particularly given that the matrix, as an entire entity, resides in a vector space.

These limitations motivate us to revisit the definition of direction and magnitude in the context of matrix-based adaptation. 
Rather than interpreting direction at the column level, we propose to reformulate weight matrices as vectors in a high-dimensional vector space through \textit{flattening}.
Specifically, a matrix $\mathbf{W} \in \mathbb{R}^{n \times m}$ can be vectorized into a vector $\mathbf{w} \in \mathbb{R}^{nm}$.
This transformation enables a more principled interpretation of direction and magnitude, leveraging well-established concepts from vector calculus. 
Under this formulation, the direction of $\mathbf{w}$ is given by its normalized vector, and the magnitude corresponds to its $\ell_2$ norm (i.e., Frobenius norm of the original matrix).

Building on this perspective, we propose a novel framework, MAP, which optimizes the direction and to enables the mapping of pre-trained weights into task-specific representations.
Specifically, MAP first normalizes the flattened pre-trained weight vector $\mathbf{w}$ and then learns a directional update vector $\Delta \mathbf{w}$ to adjust its orientation in the parameter space. 
To further enhance flexibility, MAP introduces two learnable scalar coefficients that independently control the magnitudes of the normalized pre-trained vector $\mathbf{w}$ and the update vector $\Delta \mathbf{w}$, without altering their respective directions.
By decoupling and learning both direction and magnitude, MAP facilitates more precise and interpretable task-specific tuning. 
Furthermore, since most existing PEFT methods—such as LoRA—focus on modeling $\Delta \mathbf{w}$, MAP is highly modular and can be readily integrated into these frameworks as a drop-in enhancement.
Extensive experiments across diverse benchmarks demonstrate that MAP can consistently improve downstream performance when coupling with existing methods.

\section{Related Work}

\subsection{Parameter Efficient Fine-tuning}
To mitigate the computational overhead of adapting large-scale models, parameter-efficient fine-tuning (PEFT) has gained prominence as a practical alternative to full model tuning. Current PEFT methodologies can be broadly classified into three paradigms \cite{ding2023parameter}: adapter-based techniques \cite{zhang2022adaptive,chen2022adaptformer,pfeiffer2020adapterfusion,he2021towards}, prefix-based approaches \cite{li2021prefix,fischer2024prompt,liu2023gpt,lester2021power,razdaibiedina2023residual,shi2023dept}, and low-rank adaptation methods \cite{hu2021lora,hyeon2021fedpara,liu2024dora,qiu2023controlling,renduchintala2023tied,kopiczko2023vera,yeh2023navigating,zhang2022adaptive}.
Adapter-based methods augment neural networks by inserting lightweight modules either sequentially or in parallel with existing layers.
These compact components enable task-specific adjustments while maintaining the integrity of the original architecture. 
Prefix-based strategies, on the other hand, prepend trainable embeddings, often termed soft prompts, to the model's input space. 
By optimizing these task-specific embeddings, the model's behavior can be steered without modifying its core parameters.
The third category, pioneered by LoRA, reparameterizes weight updates through low-rank decomposition. 
This approach approximates the update matrix as a product of two smaller matrices, significantly reducing the number of trainable parameters while preserving adaptation capacity.

\subsection{Low-rank Adaptation}
LoRA leverages the observation that weight updates during fine-tuning often have a low intrinsic rank, allowing task-specific adaptation to be captured by a low-rank approximation \cite{aghajanyan2020intrinsic, li2018measuring}. 
For a pre-trained weight matrix $\mathbf{W} \in \mathbb{R}^{n \times m}$, LoRA introduces a low-rank update $\Delta \mathbf{W} =\mathbf{A} \mathbf{B}$, where $\mathbf{A} \in \mathbb{R}^{n \times r}$ and $\mathbf{B} \in \mathbb{R}^{r \times m}$  with the rank $r \ll \{n,m\}$. 
During fine-tuning, only $\mathbf{A}$ and $\mathbf{B}$ are updated, while $\mathbf{W}$ remains frozen. 
The final fine-tuned weights are given by:
\begin{equation}
\mathbf{W} \rightarrow \mathbf{W} + \Delta \mathbf{W} = \mathbf{W} + \mathbf{A}\mathbf{B}.
\label{eq lora}
\end{equation}
At initialization, the matrix $\mathbf{A}$ is typically initialized using a Kaiming distribution \cite{he2015delving}, and $\mathbf{B}$ is initialized to zeros. 
During inference, the low-rank matrices $\mathbf{A}$ and $\mathbf{B}$ are integrated into $\mathbf{W}$ without any additional computational overhead. 

\subsection{Advancement in Low-rank Adaptation}
Since its inception, LoRA has inspired numerous extensions that refine its core principles \cite{hyeon2021fedpara,liu2024dora,zhang2022adaptive,si2024flora,feng2024trilora,kopiczko2023vera}.
AdaLoRA \cite{zhang2022adaptive} enhances parameter efficiency by applying singular value decomposition to weight updates, selectively retaining only the most significant components. 
FLoRA \cite{si2024flora} introduces a Tucker decomposition-based framework, constructing a low-rank core space that facilitates efficient weight reconstruction. 
In the domain of diffusion models, OFT \cite{qiu2023controlling} demonstrates the effectiveness of orthogonal transformations for parameter-efficient adaptation.
Our work builds upon these advancements in low-rank adaptation, proposing a novel approach that can intergrade with any LoRA variants. 
Through comprehensive empirical evaluation, we demonstrate the efficacy of our technique relative to state-of-the-art LoRA variants.

\subsection{Weigh Decomposed Low-rank Adaptation}
DoRA \cite{liu2024dora} extends the standard LoRA paradigm by decoupling the magnitude and direction of weights.
Specifically, DoRA proposes to decouple the adaptation process by normalizing the sum of the pre-trained weight and low-rank update $\mathbf{W} + \mathbf{AB}$ on a per-column basis, followed by rescaling each column with a learnable vector $\mathbf{m} \in \mathbb{R}^m$. The final weight is computed as:
\begin{equation}
\text{DoRA} = \mathbf{m} \cdot \frac{\mathbf{W} + \mathbf{AB}}{\|\mathbf{W} + \mathbf{AB}\|_c},
\end{equation}
where $\|\cdot\|_c$ denotes column-wise normalization.
Beyond DoRA, BiDoRA \cite{qin2024bidora} introduces a bi-level optimization scheme to decouple the learning of magnitude and direction in DoRA-style adaptation. 
BoRA \cite{wang2024bora} extends the decomposition strategy of DoRA by introducing symmetric modulation across both row and column dimensions, addressing DoRA’s vertical-only adaptation and achieving improved alignment in weight structure and downstream performance.

While both MAP and DoRA share the high-level concept of the magnitude and direction of weight updates in parameter-efficient fine-tuning, their formulations, motivations, and implementations are fundamentally different.
Specifically, 
\begin{itemize}
    \item DoRA implicitly assumes that the direction of a matrix can be decomposed into per-column units, an assumption that lacks a clear theoretical grounding in matrix analysis. 
    In contrast, MAP revisits the definition of direction and magnitude from a vector space perspective.
    Instead of defining them column-wise, we flatten the entire matrix into a vector and apply standard vector normalization.
    This formulation respects the global structure of the matrix and avoids column-wise decomposition. 
    \item DoRA introduces $m$ additional learnable parameters for each $n \times m$ weight matrix, which can not be negligible. However, MAP introduces only two additional parameters per matrix, making it significantly more parameter-efficient than DoRA. 
    Despite using fewer parameters, our method achieves better performance than DoRA in the experimental results.
    \item DoRA is a method, while MAP is a framework, which can be coupled with existing methods to enhance their performances.
\end{itemize}

These differences highlight that, although both methods conceptually mention ``direction'' and ``magnitude'', MAP and DoRA are largely unrelated in terms of both theoretical motivation and practical behavior.

\section{The Proposed Framework: MAP}

\begin{figure*}
    \centering
    \includegraphics[width=\textwidth]{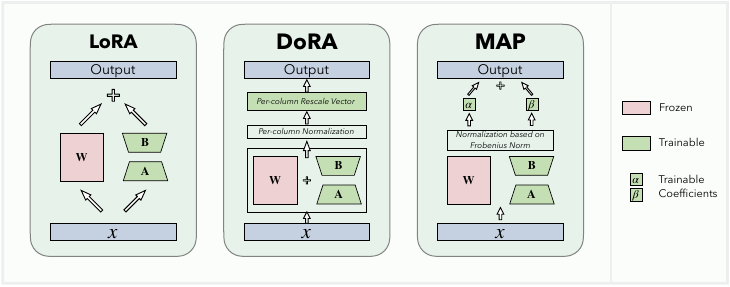}
    \caption{Comparison of LoRA, DoRA, and our proposed MAP framework. DoRA normalizes the sum $\mathbf{W} + \mathbf{A}\mathbf{B}$ column-wise and rescales each column using a trainable vector.
    In contrast, MAP normalizes both $\mathbf{W}$ and $\mathbf{A}\mathbf{B}$ using their Frobenius norms, and applies two learnable scalar coefficients $\alpha$ and $\beta$ to decouple and modulate their magnitudes. MAP provides a more principled and compact decoupling strategy in the vector space.}
    \label{fig:framework}
\end{figure*}

In this section, we introduce our framework, MAP.
The framework of MAP is shown in Fig. \ref{fig:framework}.
We first revisit the notion of matrix adaptation from a vector space perspective.
As discussed in the introduction, any weight matrix $\mathbf{W} \in \mathbb{R}^{n \times m}$ can be flattened into a high-dimensional vector $\mathbf{w} \in \mathbb{R}^{nm}$. 
This allows us to interpret the adaptation process in terms of classical vector operations, where the direction of $\mathbf{w}$ is defined by its unit-norm vector, and the magnitude by its $\ell_2$ norm.

\subsection{Vector-Based Formulation}

Under this formulation, we model the final fine-tuned weight vector $\mathbf{w}'$ as a directional combination of the pre-trained weights and the learned update vector:
\begin{equation}
\mathbf{w}' = \alpha \cdot \frac{\mathbf{w}}{\|\mathbf{w}\|} + \beta \cdot \frac{\Delta\mathbf{w}}{\|\Delta\mathbf{w}\|},
\label{eq:vector-map}
\end{equation}
where $\alpha, \beta \in \mathbb{R}$ are learnable scalar coefficients that independently control the contribution (i.e., magnitude) of the pre-trained vector and its directional update.
This formulation has several appealing properties:
\begin{itemize}
    \item It cleanly separates the direction and magnitude of each component.
    \item It treats $\mathbf{w}$ and $\Delta \mathbf{w}$ in the same vector space, enabling their alignment or contrast to be interpreted geometrically.
    \item It introduces only two additional parameters per layer, making it highly lightweight compared to methods like DoRA
\end{itemize}

\subsection{Matrix-Based Formulation}

When applied to the original matrix space, particularly when integrated with LoRA, MAP takes the following form:
\begin{equation}
\mathbf{W}^* = \alpha \cdot \frac{\mathbf{W}}{\|\mathbf{W}\|_F} + \beta \cdot \frac{\mathbf{A}\mathbf{B}}{\|\mathbf{A}\mathbf{B}\|_F},
\label{eq:matrix-map}
\end{equation}
where $\|\cdot\|_F$ denotes the Frobenius norm, and $\Delta \mathbf{W} = \mathbf{A}\mathbf{B}$ is the standard low-rank adaptation used in LoRA.
This formulation can be directly derived since the Frobenius norm of $\mathbf{W}$ satisfies:
\begin{equation}
\|\mathbf{W}\|_F = \|\mathbf{w}\|_2,
\end{equation}
Therefore, the vector and matrix formulations of MAP are mathematically equivalent in terms of magnitude scaling.

\begin{table*}[!ht]
 \renewcommand\arraystretch{1}
    \centering
    \caption{Results on commonsense reasoning tasks. Results of all the baseline methods are taken from \cite{si2025unleashing, wu2024mixture}.}
    \resizebox{\textwidth}{!}{
    \begin{tabular}{c c | c c c  c c c c c | l }
    \toprule
   \textbf{Method} & \textbf{Params(\%)} & \textbf{BoolQ} & \textbf{PIQA} & \textbf{SIQA} & \textbf{HellaS.} & \textbf{WinoG.} & \textbf{ARC-e} & \textbf{ARC-c} & \textbf{OBQA} & \multicolumn{1}{c}{\textbf{Avg.}} \\
    \toprule
    ChatGPT & - & 73.1 & 85.4 & 68.5 & 78.5 & 66.1 & 89.8 & 79.9 & 74.8 & 77.0 \\ \midrule
    
    \multicolumn{11}{c}{\textit{Fine-tuning LLaMA-7B}} \\ \midrule
    
    Fully FT & 100\% & 69.9 & 84.2 & 78.9 & 92.3 & 83.3 & 86.6 & 72.8 & 83.4 & 81.4 \\ \midrule

    Prefix & 0.11\% & 64.3 & 76.8 & 73.9 & 42.1 & 72.1 & 72.9 & 54.0 & 60.6 & 64.6 \\

    Series & 0.99\% & 63.0 & 79.2 & 76.3 & 67.9 & 75.7 & 74.5 & 57.1 & 72.4 & 70.8 \\

    Parallel & 3.54\% & 67.9 & 76.4 & 78.8 & 69.8 & 78.9 & 73.7 & 57.3 & 75.2 & 72.2 \\
    
    LoRA$_{r=4}$ & 0.10\% & 2.3 & 46.1 & 18.3 & 19.7 & 55.2 & 65.4 & 51.9 & 57.0 & 39.5 \\  
    
    AdaLoRA$_{r=4}$ & 0.10\% & 66.1 & 78.1 & 74.3 & 34.0 & 74.4 & 76.7 & 57.5 & 71.2 & 66.5 \\

    FLoRA$_{r=4}$ & 0.10\% & \textbf{}67.2 & 78.0 & 72.9 & 65.4 & 73.8 & 73.8 & 55.3 & 71.8 & 69.8 \\

    DoRA$_{r=4}$ & 0.10\% & 51.3 & 42.2 & 77.8 & 25.4 & 78.8 & 78.7 & 62.5 & 78.6 & 61.9 \\
    
    \rowcolor{gray!20}
    
    LoMAP & 0.10\% & 69.3 & 78.4 & 76.3 & 83.4 & 81.0 & 78.2 & 63.1 & 77.2 & 75.9 \\  \midrule

    LoRA$_{r=8}$ & 0.21\% & 31.3 & 57.0 & 44.0 & 11.8 & 43.3 & 45.7 & 39.2 & 53.8 & 40.7 \\ 

    \rowcolor{gray!20}

    LoMAP & 0.21\% & 69.3 & 80.6 & 78.5 & 84.0 & 79.5 & 79.0 & 63.1 & 77.2 & 76.4 \\ \midrule

    LoRA$_{r=16}$ & 0.42\% & 69.9 & 77.8 & 75.1 & 72.1 & 55.8 & 77.1 & 62.2 & 78.0 & 70.9 \\ \rowcolor{gray!20}
    
    LoMAP & 0.42\% & 69.6 & 81.6 & 78.3 & 85.1 & 81.5 &  81.3 & 66.7 & 78.8 & 77.9 \\ 

    \midrule
     
    LoRA$_{r=32}$ & 0.83\% & 68.9 & 80.7 & 77.4 & 78.1 & 78.8 & 77.8 & 61.3 & 74.8 & 74.7 \\  
    
    AdaLoRA$_{r=32}$ & 0.83\% & 69.1 & 82.2 & 77.2 & 78.3 & 78.2 & 79.7 & 61.9 & 77.2 & 75.5 \\

    FLoRA$_{r=32}$ & 0.83\% & 66.4 & 81.3 & 77.1 & 75.6 & 77.1 & 77.2 & 62.4 & 77.6 & 74.3  \\ 

    DoRA$_{r=32}$ & 0.84\% & 69.7 & 83.4 & 78.6 & 87.2 & 81.0 & 81.9 & 66.2 & 79.2 & 78.4 \\

    LoRA-Dash$_{r=32}$ & 0.83\% & 69.9 & 82.8 & 78.6 & 84.9 & 81.6 & 82.3 & 66.5 & 80.8 & 78.4 \\

    \rowcolor{gray!20}

     LoMAP & 0.83\% & 69.0 & 82.7 & 78.2 & 87.9 & 82.2 &  83.3 & 65.9 & 81.0 & 78.8 \\

    \midrule

    \multicolumn{11}{c}{\textit{Fine-tuning LLaMA3-8B}} \\ \midrule

    Fully FT & 100\% & 75.3 & 89.9 & 81.5 & 95.8 & 87.6 & 91.6 & 79.3 & 87.4 & 86.1 \\
    
    \midrule
    
    LoRA$_{r=16}$ & 0.35\% & 72.3 & 86.7 & 79.3 & 93.5 & 84.8 & 87.7 & 75.7 & 82.8 & 82.8 \\ 
    
    AdaLoRA$_{r=16}$ & 0.35\% & 73.0 & 86.7 & 77.6 & 83.3 & 83.4 & 90.2 & 78.6 & 84.2 & 82.1 \\

    FLoRA$_{r=16}$ & 0.35\% & 73.1 & 86.7 & 77.9 & 91.3 & 83.9 & 88.8 & 77.1 & 80.5 & 82.4 \\

    \rowcolor{gray!20}

     LoMAP & 0.35\% & 74.3 & 89.0 & 80.5 & 95.1 & 87.0 & 90.1 & 79.9 & 85.6 & 85.2 \\

    \midrule

    LoRA$_{r=32}$ & 0.70\% & 70.8 & 85.2 & 79.9 & 91.7 & 84.3 & 84.2 & 71.2 & 79.0 & 80.8 \\ 
    
    PISSA$_{r=32}$ & 0.70\% & 67.1 & 81.1 & 77.2 & 83.6 & 78.9 & 77.7 & 63.2 & 74.6 & 75.4\\

    MiLoRA$_{r=32}$ & 0.70\% & 68.8 & 86.7 & 77.2 & 92.9 & 85.6 & 86.8 & 75.5 & 81.8 & 81.9 \\

    DoRA$_{r=32}$ & 0.71\% & 74.6 & 89.3 & 79.9 & 95.5 & 85.6 & 90.5 & 80.4 & 85.8 & 85.2 \\
    
    \rowcolor{gray!20}

    LoMAP & 0.70\% & 75.7 & 88.4 & 79.8 & 95.5 & 87.3 & 90.8 & 81.5 & 88.0 & 85.8 \\
    
    \bottomrule
    \end{tabular}}
    \label{tab:results of commonsense}
\end{table*}

In this way, MAP can be viewed as a direction-aware modulation layer that scales the normalized base and update matrices in a principled and learnable fashion.
We advocate LoMAP as a drop-in enhancement to existing PEFT frameworks.
In our experiments, we integrate MAP with LoRA to form a new variant, LoMAP, which we adopt as the default implementation.

\section{Experiments}

In this section, we conduct a series of experiments to demonstrate the effectiveness of MAP across various tasks, including commonsense reasoning, natural language understanding, and subject-driven generation tasks.
In the following subsections, we provide detailed descriptions of each task and report the corresponding performance achieved by MAP.
The parameters are initialized with $\alpha=\|\mathbf{W}\|_F$ and $\beta=1$ for all tasks.

\begin{table*}[ht]
    \centering
    \renewcommand\arraystretch{1} 
    \caption {Results with DeBERTaV3 fine-tuned on GLUE development set. ``FT'' represents fully fine-tuning, and ``Base' and ``Large'' represent DeBERTaV3-base and DeBERTaV3-large, respectively.}
    \resizebox{\textwidth}{!}{
    \begin{tabular}{l | c| c c c c c c c c |>{\columncolor{gray!10}}c}
    \toprule
         \multirow{2}{*}{\textbf{Method}} &  \multirow{2}{*}{\textbf{Params(\%)}} & \textbf{MNLI} & \textbf{SST-2} &\textbf{CoLA} & \textbf{QQP} & \textbf{QNLI} & \textbf{RTE} & \textbf{MRPC} & \textbf{STS-B} & \textbf{All}\\
         & & Acc & Acc & Mcc & Acc & Acc & Acc & Acc & Corr & Avg. \\ 
         \midrule
        
        Base(FT) & 100\% & 89.90 & 95.63 & 69.19 & 91.87 & 94.03 & 83.75 & 90.20 & 91.60 & 88.27 \\ \midrule

        Series & 0.17\% & 90.10 & 95.41 & 67.65 & 91.19 & 93.52 & 83.39 & 89.25 & 91.31 & 87.73 \\
        
        Padapter & 0.16\% & 89.89 & 94.72 & 69.06 & 91.05 & 93.87 & 84.48 & 89.71 & 91.38 & 88.02 \\
        
         LoRA$_{r=2}$ & 0.18\% & 90.03 & 93.92 & 69.15 & 90.61 & 93.37 &  87.01 & 90.19 & 90.75 & 88.13  \\

         DoRA & 0.22\% & 90.21 & 94.38 & 69.33 & 90.84 & 93.26 & 86.94 & 90.19 & 91.34 & 88.31 \\

         \rowcolor{gray!20}

         LoMAP & 0.18\% & 90.52 & 95.91 & 70.38 & 91.83 & 94.31 & 89.16 & 91.67 & 92.14 & 89.49\\  \midrule

        LoRA$_{r=8}$ & 0.72\% & 89.80 & 93.69 & 69.30 & 91.78 & 92.97 & 86.28 & 90.68 & 91.62 & 88.27  \\
        
        DoRA & 0.77\% & 89.67 & 94.61 & 69.08 & 91.80 & 93.23 & 87.33 & 90.68 & 91.73 & 88.49 \\
        
        \rowcolor{gray!20}

         LoMAP & 0.72\% & 90.71 & 96.13 & 71.08 & 92.19 & 94.53 & 89.07 & 91.67 & 91.76 & 89.64  \\  \bottomrule \toprule

         Large(FT) & 100\% & 91.81 & 96.93 & 75.27 & 93.01 & 96.02 & 92.68 & 92.20 & 92.98 & 91.36 \\ \midrule
        
         LoRA$_{r=2}$ & 0.20\% & 91.33 & 95.87 & 73.89 & 91.84 & 95.14 & 91.69 & 90.68 & 92.85 & 90.41 \\  \rowcolor{gray!20}

         LoMAP & 0.20\% & 91.82 & 96.52 & 74.31 & 92.23 & 95.58 & 92.43 & 92.75 & 92.89 & 91.07 \\  \midrule

        LoRA$_{r=8}$ & 0.80\% & 91.38 & 96.33 & 74.48 & 92.54 & 95.48 & 92.05 & 91.17 & 92.92 & 90.79 \\  \rowcolor{gray!20}

         LoMAP & 0.80\% & 91.72 & 96.39 & 75.21 & 92.82 & 95.82 & 92.78 & 91.69 & 93.02 & 91.18 \\  
         
        \bottomrule
    \end{tabular}
    }
    \label{tab: deberta results}
\end{table*}

\subsection{Commonsense Reasoning}

\subsubsection{Task, Model, and Baselines}

The commonsense reasoning evaluation includes eight diverse benchmarks, each associated with a specific dataset: BoolQ \cite{clark2019boolq}, PIQA \cite{bisk2020piqa}, SIQA \cite{sap2019socialiqa}, HellaSwag \cite{zellers2019hellaswag}, WinoGrande \cite{sakaguchi2021winogrande}, ARC-e, ARC-c \cite{clark2018thinkarce}, and OpenBookQA (OBQA) \cite{mihaylov2018canobqa}. 
Following the protocol proposed in \cite{hu2023llm}, we consolidate the training splits of all benchmarks into a unified dataset, referred to as Commonsense170K, and evaluate model performance on the test sets of each benchmark individually.
We fine-tune LLaMA-7B \cite{touvron2023llama} and LLaMA3-8B \cite{llama3modelcard} on this target task. 

We compare LoMAP with several baselines including Prefix \cite{li2021prefix}, Series \cite{houlsby2019parameter}, Parallel \cite{he2021towards}, LoRA \cite{hu2021lora}, AdaLoRA \cite{zhang2022adaptive}, FLoRA \cite{si2024flora}, DoRA \cite{liu2024dora}, PISSA \cite{meng2024pissa}, and MiLoRA \cite{wang2024milora}.
In addition, we include comparisons with ChatGPT (gpt-3.5-turbo) by leveraging its zero-shot Chain-of-Thought reasoning capabilities, as outlined in \cite{wei2022chain}.
All the experiments are conducted using NVIDIA A100 GPUs. 
The hyper-parameters are
shown in Table \ref{tab: cr detail}.

\subsubsection{Experimental Results}

Table \ref{tab:results of commonsense} presents the evaluation results of various PEFT methods on commonsense reasoning benchmarks under multiple model backbones and rank settings. 
We observe several notable trends:

First, LoMAP consistently outperforms all competing PEFT baselines under similar parameter budgets. 
For instance, under the LLaMA-7B backbone with $r=16$, LoMAP achieves an average accuracy of 77.9, surpassing both DoRA (78.4) and LoRA (70.9).
Notably, even at lower ranks (e.g., $r=4$ or $r=8$), LoMAP yields strong performance, achieving 75.9 at $r=4$ and 76.4 at $r=8$, clearly outperforming AdaLoRA, FLoRA, and DoRA at the same ranks.
Second, LoMAP demonstrates excellent scalability across model sizes. 
When evaluated on the larger LLaMA3-8B model, LoMAP achieves the highest accuracy of 85.8 at $r=32$, surpassing strong baselines such as DoRA (85.2), and AdaLoRA (82.1). 
This indicates that LoMAP remains effective even when scaling to larger models and more complex reasoning tasks.
These results collectively validate the effectiveness and generality of LoMAP as a principled and scalable improvement over existing low-rank adaptation methods.

\subsection{Natural Language Understanding}

\subsubsection{Task, Model, and Baselines}

For the Natural Language Understanding (NLU) task, we use the General Language Understanding Evaluation (GLUE) benchmark \cite{wang2018glue}, which evaluates models across various tasks. 
The benchmark includes two sentence classification tasks: CoLA \cite{warstadt2019neural} and SST-2 \cite{socher2013recursive}, three tasks related to similarity and paraphrasing: MRPC \cite{dolan2005automatically}, QQP \cite{wang2018glue}, and STS-B \cite{cer2017semeval}, as well as three natural language inference tasks: MNLI \cite{williams2017broad}, QNLI \cite{rajpurkar2016squad}, and RTE \cite{dagan2005pascal,bar2006second,giampiccolo2007third,bentivogli2009fifth}. Detailed information about these datasets is provided in Table \ref{tab: glue dataset}.
We fine-tune the DeBERTaV3-base and DeBERTaV3-large \cite{he2021debertav3} models on these tasks. 
The hyper-parameter settings are shown in Table \ref{tab: nlu detail}.

In addition to LoRA, DoRA and Series methods, we also include PAdapter in our comparisons.
Series introduces adapter modules at the interface between the self-attention and FFN blocks, incorporating them with residual connections to preserve model flow. In contrast, PAdapter employs a more streamlined design by attaching adapters solely after the FFN and LayerNorm layers.

\subsubsection{Experimental Results}

Table \ref{tab: deberta results} presents the GLUE benchmark results with DeBERTaV3-base and DeBERTaV3-large under various PEFT settings. 
Across both model sizes and rank configurations, LoMAP consistently delivers superior performance.

Under the DeBERTaV3-base setting, LoMAP exhibits clear advantages even in low-rank regimes.
At rank $r=2$, it achieves an average score of 89.49, outperforming LoRA and Padapter, while maintaining comparable parameter efficiency. 
As the rank increases to $r{=}8$, LoMAP continues to lead, reaching 89.64, surpassing LoRA’s 88.27 by a substantial margin.
For the larger DeBERTaV3-large model, the benefits of LoMAP remain prominent. 
With $r{=}8$, it achieves 91.18 average score, closing the gap with full fine-tuning (91.36) while requiring less than 1\% of the trainable parameters. 
This demonstrates LoMAP’s strong capacity to scale to more expressive architectures without sacrificing efficiency.
These results confirm the practicality of LoMAP for general-purpose language understanding and its potential to replace conventional fine-tuning in resource-constrained settings.

\subsection{Subject-driven Generation}

\begin{figure*}[ht]
    \centering
    \includegraphics[width=\textwidth]{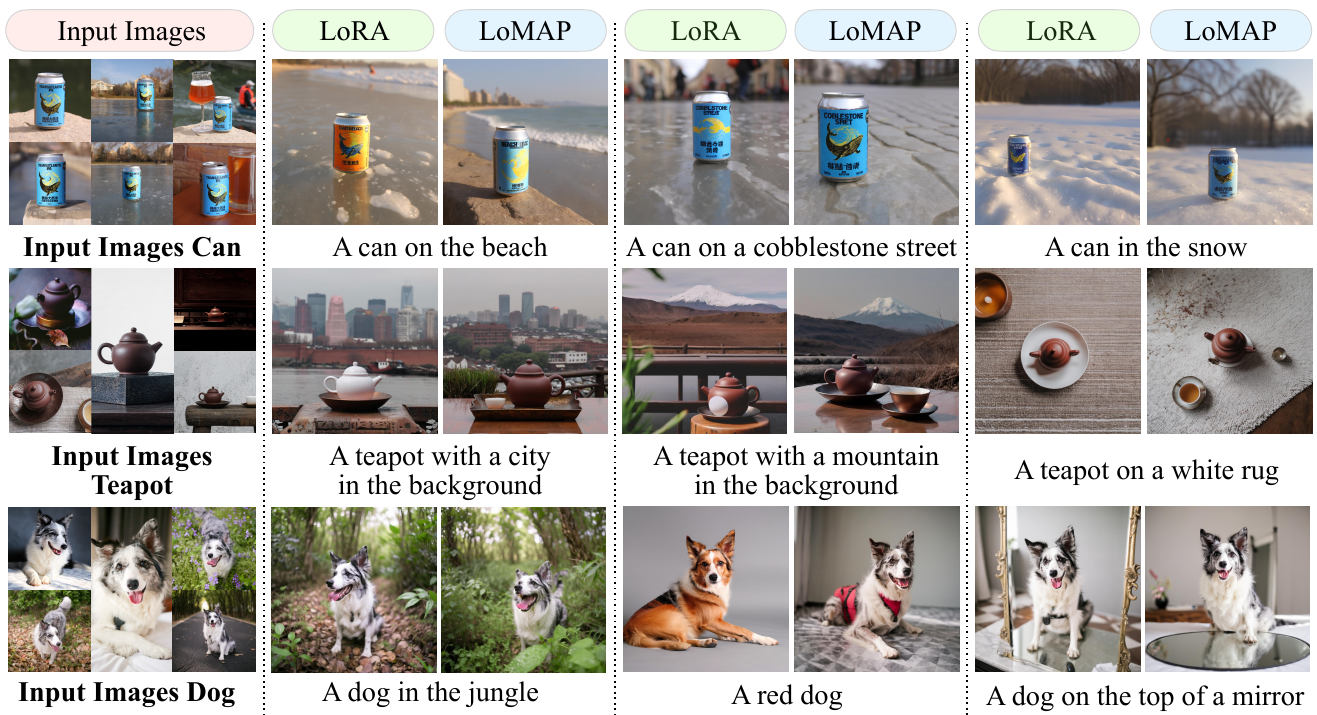}
    \caption{Comparison of generated images from LoRA and LoMAP on the subject-driven generation task. It is evident that LoMAP consistently produces images that better reflect both the input subjects and the intended prompts compared to standard LoRA.}
    \label{fig: sdg}
\end{figure*}

\begin{table*}[ht]
    \centering
    \caption{Comparison of joint versus stepwise optimization on LLaMA3-8B.}
    \renewcommand\arraystretch{1}

    \resizebox{\textwidth}{!}{
    \begin{tabular}{l c|cccccccc|c}
    \toprule
    \textbf{Method} & \textbf{Params (\%)} & \textbf{BoolQ} & \textbf{PIQA} & \textbf{SIQA} & \textbf{HellaS.} & \textbf{WinoG.} & \textbf{ARC-e} & \textbf{ARC-c} & \textbf{OBQA} & \textbf{Avg.} \\
    \midrule
    Joint & 0.70 & 75.7 & 88.4 & 79.8 & 95.5 & 87.3 & 90.8 & 81.5 & 88.0 & 85.8 \\
    
    Stepwise & 0.70 & 75.1 & 88.7 & 80.1 & 95.3 & 87.4 & 90.4 & 81.2 & 88.3 & 85.8 \\
    \bottomrule
    \end{tabular}}
    \label{tab:stepwise_vs_joint}
\end{table*}

\subsubsection{Task, Model, and Baselines}

In this experiment, we fine-tune text-to-image diffusion models for subject-driven image generation, following the setup proposed in DreamBooth \cite{ruiz2023dreambooth}.
The goal is to synthesize images that faithfully reflect a specific subject, given only a few reference examples. 
To achieve this, we fine-tune a text-to-image model using image-text pairs in which the subject is denoted by a unique identifier (e.g., ``A photo of a [V] cat'').
After fine-tuning, this identifier is embedded into new prompts to guide image generation specific to the learned subject.

We use the SDXL5 model \cite{podell2023sdxl} as the backbone and apply both LoRA and LoMAP as  techniques. 
The model is trained with a learning rate of 1e-4, a batch size of 4, and for 500 steps on a single 80GB A100 GPU, which takes approximately 26 minutes. 
Image generation is conducted using 50 inference steps per prompt, with each synthesis taking around 10 seconds.
All experiments are conducted using the official DreamBooth dataset \cite{ruiz2023dreambooth}.

\subsubsection{Experimental Results}

As illustrated in Fig. \ref{fig: sdg}, the qualitative results highlight that LoMAP yields images with greater subject fidelity compared to standard LoRA. 
In particular, LoRA’s generated samples—for example, those depicting a dog or a teapot—often diverge noticeably from the reference images.
In contrast, LoMAP consistently preserves key subject attributes, producing visuals more closely aligned with the original exemplars.
In addition, LoMAP demonstrates strong semantic alignment with complex prompts, accurately interpreting and visually rendering fine-grained concepts such as \textit{cobblestone} or \textit{white rug}.
This highlights LoMAP’s capacity to effectively disentangle and integrate subject- and prompt-specific information during synthesis.

\begin{table*}[!ht]
 \renewcommand\arraystretch{1}
    \centering
    \caption{Results of AdaLoRA and FLoRA with MAP on LLaMA3-8B for commonsense reasoning tasks.}
    \resizebox{\textwidth}{!}{
    \begin{tabular}{c c | c c c c c c c c |>{\columncolor{gray!10}}c }
    \toprule
    \textbf{Method} & \# \textbf{Params (\%)} & \textbf{BoolQ} & \textbf{PIQA} & \textbf{SIQA} & \textbf{HellaSwag} & \textbf{WinoGrande} & \textbf{ARC-e} & \textbf{ARC-c} & \textbf{OBQA} & \textbf{Avg.} \\
    \toprule

    AdaLoRA$_{r=16}$ & 0.35 & 73.0 & 86.7 & 77.6 & 83.3 & 83.4 & 90.2 & 78.6 & 84.2 & 82.1 \\ \rowcolor{gray!20}
    
    AdaLoMAP & 0.35 & 73.2 & 87.6 & 78.7 & 94.6 & 84.8 & 89.7 & 78.9 & 85.0 & 84.1 \\

     AdaLoRA$_{r=32}$ & 0.70 & 73.5 & 87.2 & 78.2 & 83.4 & 84.1 & 90.4 & 79.1 & 85.0 & 82.6   \\  \rowcolor{gray!20}
    
    AdaLoMAP & 0.70 & 73.9 & 87.9 & 79.9 & 95.1 & 84.8 & 89.9 & 78.8 & 85.2 & 84.4 \\
    
    \midrule
    
    FLoRA$_{r=16}$ & 0.35 & 73.1 & 86.7 & 77.9 & 91.3 & 83.9 & 88.8 & 77.1 & 80.5 & 82.4 \\

    \rowcolor{gray!20}
    
    FLoMAP & 0.35 & 74.1 & 87.7 & 80.0 & 94.6 & 84.2 & 89.8 & 78.2 & 84.0 & 84.1 \\

    FLoRA$_{r=32}$ & 0.70 & 73.3 & 87.2 & 79.5 & 93.7 & 84.8 & 88.6 & 76.4 & 84.1 & 83.5 \\  \rowcolor{gray!20}
    
    FLoMAP & 0.70 & 74.6 & 88.4 & 80.3 & 95.0 & 84.5 & 90.1 & 78.4 & 84.6 & 84.5 \\ 
    
    \bottomrule
    \end{tabular}}
    \label{tab: flora and adalora of commonsense}
\end{table*}

\section{Further Analysis}

In this section, we conduct a more in-depth investigation of MAP to further substantiate its effectiveness and clarify the underlying mechanisms contributing to its superior performance.

\subsection{Independent Optimization of Direction and Magnitude}

We consider separately optimizing the direction and magnitude components, i.e, step-wise  optimization of ($\alpha$, $\beta$) and $\Delta\mathbf{W}$. 
To evaluate this idea, we conduct experiments on LLaMA3-8B, and the results are summarized in Table \ref{tab:stepwise_vs_joint}. 
The findings suggest that there is no significant performance difference between jointly optimizing both components and optimizing their distributions independently.
However, it is important to note that performing direction and magnitude optimization in a stepwise manner introduces additional training time and breaks the standard end-to-end optimization pipeline.
Therefore, we adopt the joint optimization strategy in all our experiments for its simplicity, efficiency, and compatibility with mainstream training frameworks.

\subsection{Coupling with Other Methods}
Table~\ref{tab: flora and adalora of commonsense} presents the evaluation results of integrating MAP with two representative PEFT baselines, AdaLoRA and FLoRA, on commonsense reasoning tasks using the LLaMA3-8B backbone. 
Across all rank settings and benchmarks, we observe consistent improvements in performance when MAP is applied.
For instance, AdaLoMAP outperforms AdaLoRA at both $r=16$ and $r=32$, showing clear gains on tasks such as SIQA, HellaSwag, and WinoGrande. 
Similarly, FLoMAP demonstrates notable improvements over FLoRA, achieving an average score of 84.5 at $r=32$, compared to 83.5 from its base variant. 
The improvements observed indicate that MAP can seamlessly integrate with various PEFT methods and enhance their performance, which suggests that MAP can serve as a universal plugin.

\subsection{Training Costs}

\begin{table}[ht]
    \centering
    \setlength{\tabcolsep}{2mm}
    \renewcommand{\arraystretch}{1}
    \caption{Training time (minutes/epoch) and GPU memory usage (GB) for different methods on representative GLUE tasks using DeBERTaV3-base.}
    \resizebox{\linewidth}{!}{
    \begin{tabular}{l|cc|cc|cc}
    \toprule
    \multirow{2}{*}{\textbf{Method}} & \multicolumn{2}{c|}{\textbf{MNLI}} & \multicolumn{2}{c|}{\textbf{SST-2}} & \multicolumn{2}{c}{\textbf{STS-B}}  \\
    & \textbf{Time} & \textbf{GPU} & \textbf{Time} & \textbf{GPU} & \textbf{Time} & \textbf{GPU} \\
    
    \midrule
    
    LoRA & 73.57 & 11.35 & 6.38 & 6.85 & 0.56 & 6.85 \\
    
    DoRA & 118.42 & 16.72 & 11.26 & 9.66 & 0.91 & 9.66 \\
    
    LoMAP  & 80.41 & 12.56 & 6.92 & 7.18 & 0.62 & 7.18 \\
    
    \bottomrule
    \end{tabular}
    }
\label{tab:resource-analysis}
\end{table}

We report the time and GPU resources required by LoMAP when fine-tuning the DeBERTaV3-base model, in comparison with LoRA and DoRA. 
The results are summarized in Table \ref{tab:resource-analysis}. 
It is evident that MAP incurs negligible additional cost over LoRA, requiring comparable GPU memory and training time.
In contrast, DoRA introduces significantly higher computational overhead due to its column-wise normalization and per-column scaling, leading to increased memory consumption and slower training.
Despite its lightweight nature, MAP consistently outperforms both LoRA and DoRA in downstream performance, as shown in our experiments.
This demonstrates that MAP strikes a superior balance between efficiency and effectiveness, making it a practical and scalable enhancement to existing PEFT frameworks.

\section{Conclusion}

In this work, we revisited the foundational concepts of direction and magnitude in the context of parameter-efficient fine-tuning.
Motivated by the limitations of DoRA, particularly its heuristic column-wise normalization and high parameter overhead, we proposed a principled vectorized perspective that treats matrices as high-dimensional vectors.
Building on this insight, we introduced MAP, a simple yet effective framework that decouples and learns both the direction and magnitude of weight updates.
MAP operates by normalizing the pre-trained weights and the update directions, followed by learning two scalar magnitudes to scale each component independently.
This formulation not only enables fine-grained control and interpretability but also remains computationally lightweight and introduces minimal additional parameters.
Moreover, MAP can be seamlessly integrated into existing PEFT frameworks such as LoRA, AdaLoRA, and FLoRA, consistently boosting their performance.
Extensive experiments across language understanding, commonsense reasoning, and generation tasks validate the effectiveness, efficiency, and versatility of our approach.

\section*{Limitations}
While MAP improves flexibility by decoupling magnitude and direction, it does not explicitly account for the constraints imposed on the weight vector by the low-rank structure of $\Delta \mathbf{W}$. 
These constraints may limit the expressiveness of the learned update in certain tasks.
Future work could explore incorporating additional flexibility into the low-rank structure to address this limitation.

\bibliography{custom}

\appendix

\section{Experiment Details}\label{sec appendix detail}

\subsection{Implementation Details}

We primarily evaluate MAP in combination with LoRA, testing different rank values for LoRA and other methods from the set \{2, 4, 8, 16, 32\}. 
All experiments are implemented using the publicly available PyTorch framework \cite{paszke2019pytorch}, and all training is conducted on NVIDIA A100 GPUs. 
For consistency, we fine-tune all the linear layers of the models across all experiments.

\begin{table}[!ht]

\centering
\renewcommand\arraystretch{1}
\setlength{\tabcolsep}{1mm}
\caption{Hyper-parameter settings of LoMAP on commonsense reasoning task.}
\resizebox{\linewidth}{!}{
\begin{tabular}{c | c c c c | c c} 

\toprule

\textbf{Settings} & \multicolumn{4}{c|}{LLaMA-7B} & \multicolumn{2}{c}{LLaMA3-8B}\\ 

\midrule

Rank $r$ & 4 & 8 & 16 & 32 & 16 & 32 \\ \midrule

$\alpha$ & 32 & 64 & 32 & 64 & 32 & 64 \\ \midrule

LR ($10^{-4}$) & 3 & 3 & 2 & 3 & 3 & 3 \\ \midrule

LR Scheduler & \multicolumn{6}{c}{Linear} \\ \midrule

Dropout & \multicolumn{6}{c}{0.05} \\ \midrule

Optimizer & \multicolumn{6}{c}{AdamW} \\ \midrule

Batch size & \multicolumn{6}{c}{16} \\ \midrule

Warmup Steps & \multicolumn{6}{c}{100} \\ \midrule

Epochs & \multicolumn{6}{c}{3} \\ \midrule

Where & \multicolumn{6}{c}{Q, K, V, Up, Down} \\

\bottomrule

\end{tabular}}
\label{tab: cr detail}
\end{table}

\begin{table*}[!ht]
    \centering
    \caption{Details of GLUE dataset.}

    \resizebox{\linewidth}{!}{
    \begin{tabular}{l | l  c  c  c  c  c}
    \toprule
         Dataset & Task & \# Train & \# Dev & \# Test & \# Label & Metrics \\ \midrule
         \multicolumn{7}{c}{Single-Sentence Classification} \\ \hline
         
         CoLA & Acceptability & 8.5k & 1k & 1k & 2 & Matthews corr \\ \midrule
         
         SST-2 & Sentiment & 67k & 872 & 1.8k & 2 & Accuracy \\ \midrule
         
         \multicolumn{7}{c}{Similarity and Paraphrase} \\ \midrule

         MRPC & Paraphrase & 3.7k & 408 & 1.7k & 2 & Accuracy / F1 \\ \midrule

         QQP & Paraphrase & 364k & 40k & 391k & 2 & Accuracy / F1 \\ \midrule
         
         STS-B & Similarity & 7k & 1.5k & 1.4k & 1 & Pearson/ Spearman Corr \\  \midrule

        \multicolumn{7}{c}{Natural Language Inference} \\ \midrule
          
         MNLI & NLI & 393k & 20k & 20k & 3 & Accuracy \\ \midrule
         
         QNLI & QA/NLI & 108k & 5.7k & 5.7k & 2 & Accuracy \\ \midrule

         RTE & NLI & 2.5k & 276 & 3k & 2 & Accuracy \\
        
         \bottomrule
    \end{tabular}}
    \label{tab: glue dataset}
\end{table*}

\begin{table*}
\centering
\caption{Hyper-parameter settings of LoMAP on NLU task.}
\resizebox{\linewidth}{!}{
\begin{tabular}{c | c c c c c c c c} 

\toprule

Hyper-parameter & MNLI & SST-2 & CoLA & QQP & QNLI & RTE & MRPC & STS-B\\ 

\midrule

Optimizer & \multicolumn{8}{c}{AdamW} \\ \midrule

Warmup Ratio & \multicolumn{8}{c}{0.1} \\ \midrule

LR schedule & \multicolumn{8}{c}{Linear} \\  \midrule

Rank $r$ & \multicolumn{8}{c}{2 \& 8}\\ \midrule

LoRA alpha & \multicolumn{8}{c}{4 \& 16} \\ \midrule

Max Seq. Len. & 256 & 128 & 64 & 320 & 512 & 320 & 320 & 128 \\ \midrule

Batch Size & 32 & 32 & 32 & 32 & 32 & 32 & 32 & 32 \\ \midrule

Learning Rate & 5e-4 & 8e-4 & 8e-4 & 1e-3 & 5e-4 & 1.2e-3 & 1e-3 & 5e-4 \\ \midrule

Epochs & 12 & 24 & 25 & 5 & 5 & 50 & 30 & 25  \\ 

\bottomrule

\end{tabular}}
\label{tab: nlu detail}
\end{table*}

\end{document}